\documentclass[letterpaper, 10 pt, conference]{ieeeconf} 
\IEEEoverridecommandlockouts
\usepackage{moreverb,url}
\usepackage[colorlinks,bookmarksopen,bookmarksnumbered,citecolor=red,urlcolor=red]{hyperref}
\usepackage{siunitx}
\usepackage{url}
\usepackage{svg}
\usepackage{multicol}
\usepackage{soul}
\usepackage{color}
\usepackage{flushend}
\usepackage{times}
\usepackage{graphicx}
\usepackage{lipsum}
\usepackage{setspace}
\usepackage{epsfig}
\usepackage{amsmath}
\usepackage{bm}
\usepackage{amssymb}
\usepackage{tabularx}
\usepackage{mathtools}
\usepackage{color,soul}
\usepackage{paralist}
\usepackage{tikz}
\usepackage{float}
\usepackage{pdfpages}
\usepackage{makecell}
\usepackage{array}
\usepackage{subcaption}
\usepackage{enumerate}
\usepackage{pgfgantt}
\usepackage{amsfonts}
\usepackage{textcomp}
\usepackage{multirow,diagbox}
\usepackage{threeparttable}
\usepackage{booktabs}
\usepackage{bm}
\usepackage[ruled,vlined]{algorithm2e}

\newcolumntype{L}[1]{>{\raggedright\let\newline\\\arraybackslash\hspace{0pt}}m{#1}}
\newcolumntype{C}[1]{>{\centering\let\newline\\\arraybackslash\hspace{0pt}}m{#1}}
\newcolumntype{R}[1]{>{\raggedleft\let\newline\\\arraybackslash\hspace{0pt}}m{#1}}

\newcommand{\SE}{\mathrm{SE}(3)}

\newcommand{\campose}[2]{\prescript{#1}{}{\mathbf{X}}_{#2}}
\newcommand{\objpose}[2]{\prescript{#1}{}{\mathbf{L}}_{#2}}
\newcommand{\worldf}{W}

\newcommand{\objmotion}[3]{\prescript{#1}{#2}{\mathbf{H}}_{#3}}
\newcommand{\objf}{L}
\newcommand{\camf}{X}

\newcommand{\mpoint}[2]{\prescript{#1}{}{\mathbf{m}}_{#2}}
\newcommand{\nhpoint}[2]{\prescript{#1}{}{\tilde{\mathbf{m}}}_{#2}}

\usepackage{xcolor}  

\newcommand{\etal}{\textit{et al}.~}

\def\secref#1{Section~\ref{#1}}
\def\figref#1{Fig.~\ref{#1}}
\def\tabref#1{Table~\ref{#1}}
\def\eqref#1{(\ref{#1})}
\def\algref#1{Alg.~\ref{#1}}

\title{\LARGE \bf
DynORecon: Dynamic Object Reconstruction for Navigation}

\author{Yiduo~Wang\textsuperscript{1}, Jesse Morris\textsuperscript{1}, Lan Wu\textsuperscript{2}, Teresa Vidal-Calleja\textsuperscript{2} and~Viorela~Ila\textsuperscript{1}
\thanks{This research is funded with the support of ARIA Research and the Australian Government via the Department of Industry, Science, and Resources CRC-P program (CRCPXI000007).}
\thanks{\textsuperscript{1} Yiduo Wang, Jesse Morris and Viorela Ila are with the Australian Centre For Robotics (ACFR), University of Sydney, 2006 Sydney, Australia.
{\tt \{yiduo.wang,jesse.morris,viorela.ila\}@sydney.edu.au}}
\thanks{\textsuperscript{2} Lan Wu and Teresa Vidal-Calleja are with the University Technology Sydney, 2007 Sydney, Australia.
{\tt \{lan.wu-2,teresa.vidalcalleja\}@uts.edu.au}}
}

\begin{document}

\maketitle
\thispagestyle{empty}
\pagestyle{empty}

\begin{abstract}
This paper presents DynORecon, a Dynamic Object Reconstruction system that leverages the information provided by Dynamic SLAM to simultaneously generate a volumetric map of observed moving entities while estimating free space to support navigation. By capitalising on the motion estimations provided by Dynamic SLAM, DynORecon continuously refines the representation of dynamic objects to eliminate residual artefacts from past observations and incrementally reconstructs each object, seamlessly integrating new observations to capture previously unseen structures. 
Our system is highly efficient ($\sim$20 FPS) and produces accurate ($\sim$10 cm) reconstructions of dynamic objects using simulated and real-world outdoor datasets. 


\end{abstract}


\section{Introduction}

The abundance of dynamic objects in real-world environments pose challenges to autonomous navigation systems. 
Dense reconstruction techniques are commonly employed to represent obstacles in robotic systems, 
but they often rely on the assumption that the world is static~\cite{hornung13ar_octomap, Niessner2013tog, Kahler2015tvcg_infinitam}. 
However, rigid maps are unable to accurately represent moving objects, 
as the objects observation history will leave artefacts in the map, falsely marking free space as occupied and limiting the traversable environment, making navigation inefficient and sometimes even impossible. 
To prevent this, it is essential to accurately model object movements in the reconstruction, which requires a precise understanding of object motions. 
Advancements in Dynamic Simultaneous Localisation And Mapping (SLAM) provide insight on object motions~\cite{bescos2021ral, morris2024icra, judd2024ijrr_mvo}. 
However, these systems typically rely on sparse point clouds to map the environment, which are less effective in supporting robotic navigation and exploration tasks compared to dense reconstruction methods~\cite{Funk2021ral_supereight, reijgwart2020ral_voxgraph, Wang2021ras}. 


\begin{figure}[t]
	\centering
	\includegraphics[trim={0cm 0cm 0cm 0cm},clip,width=0.9\columnwidth]{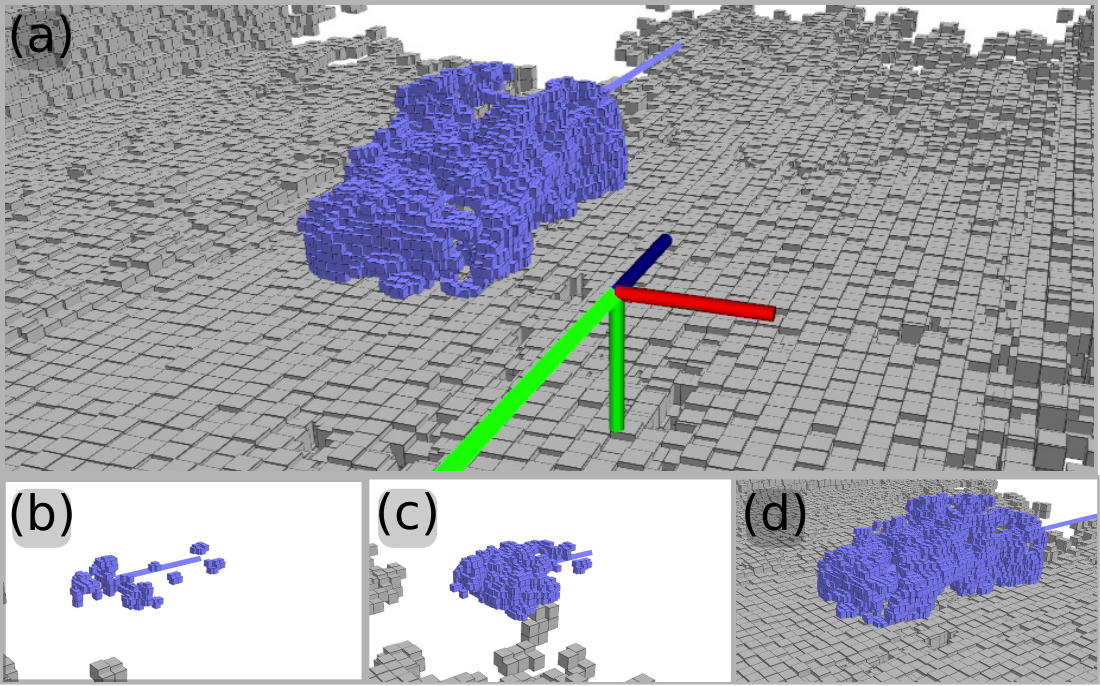}
	\caption{\small{DynORecon constructing an approaching vehicle incrementally based on a Dynamic SLAM framework. 
    \textbf{(a)}: We visualise the volumetric reconstruction of a dynamic object in addition to the object trajectory, camera pose and camera trajectory to better represent the volumes of reconstructed obstacles. We also visualise the dense reconstruction of the static environment. 
    \textbf{(b)}: Initial observation of the object. \textbf{(c--d)}: Object reconstruction being completed as more observations are integrated into it.}}
    \label{fig:dynorecon_teaser}
    \vspace{-6mm}
\end{figure}


Recent developments in both state-of-the-art dense reconstruction methods identify and address the complications of dynamic environments~\cite{Newcombe2015CVPR_dynamicfusion, Runz2018ismar_maskfusion, Schmid2023dynablox, Barad2024arxiv, Schmid2024rss_Khronos}, 
each of which focuses on one aspect of reconstruction to handle dynamics more efficiently.
On one hand, 
it is vital to accurately track the free space for the purpose of navigation and obstacle avoidance; 
on the other hand, previous observations of a dynamic object provides essential information on its total occupied volume.
Hence we desire a combination of these features: 
incorporating all previous observations to map each dynamic object without affecting the static background
while explicitly representing known free space. 



To this end, 
we propose DynORecon, 
a novel system designed to incrementally reconstruct moving objects within dynamic environments in addition to the static background by utilising an understanding of object motions in the scene provided by a state-of-the-art Dynamic SLAM system~\cite{morris2024icra}.
Our unique approach is able to completely reconstruct dynamic objects by dividing the overall reconstruction into several small-scale submaps~\cite{reijgwart2020ral_voxgraph, Wang2021ras}, one for each object.
Each submap is kept consistent with the real object by propagating its pose through both time and space using the estimated object motions from our Dynamic SLAM pipeline. 
New observations of these objects are then incrementally integrated into their corresponding models as shown in~\figref{fig:dynorecon_teaser}. 
By accumulating measurements of the object, its full volume can be reconstructed even when only partial views of the object are currently available, allowing the total occupied space of the scene to be more accurately represented and enhancing the safety and efficiency of planning algorithms.

We model obstacles in the environment using a Euclidean Signed Distance Field (ESDF) as it provides an accurate representation for surfaces and is well suited for path planning and navigation~\cite{Oleynikova2017iros_voxblox}. 
We additionally model the known free space to explicitly represent traversable regions for navigation. 
To efficiently compute ESDF, we choose VDB-GPDF~\cite{wu2024vdb_gpdf} as the core of our reconstruction as it uses Gaussian Process Distance Fields (GPDF)~\cite{wu2021faithful,legentil2023accurate} to directly estimate ESDF. 
\figref{fig:dynorecon_teaser} presents the surface voxels, and by applying Marching Cubes~\cite{lorensen1998marching}, we can obtain accurate surface meshes too. 
This framework is built upon the VDB data structure~\cite{museth2013vdb,museth2013openvdb} allowing highly efficient voxel lookup and making it suitable for real-time reconstruction of large-scale environments and navigation tasks.

The contributions of this work are as follows:
\begin{itemize}
\item A novel reconstruction method, DynORecon, for dynamic environments that can incrementally build up the full volume of each moving objects, leveraging object motion estimations. 
\item A submap structure for each dynamic object, mitigating any motion artefacts in the static map. 
\item An explicit representation of known free space to better facilitate navigation in complex dynamic environments. 
\item A thorough evaluation of DynORecon using simulation and real-world experiments, achieving 20 FPS in large-scale outdoor experiments. 
\end{itemize}
\section{Related Works}
\label{sec:related_works}

DynORecon focuses on dense volumetric reconstruction for moving objects in dynamic environments. 
As with many SLAM systems that assume the scene to be predominantly static, 
traditional reconstruction pipelines aim at mapping a static environment as one rigid model~\cite{Newcombe2011ismar, Niessner2013tog, Kahler2015tvcg_infinitam, Oleynikova2017iros_voxblox}. 
Most of these methods employ voxel maps, 
and represent the environment by storing information in each voxel such as Occupancy Probability~\cite{hornung13ar_octomap, Funk2021ral_supereight, Wang2021ras} and Signed Distance Field (SDF)~\cite{Oleynikova2017iros_voxblox, reijgwart2020ral_voxgraph, Funk2021ral_supereight}. 
To handle dynamic objects so that their history does not leave residual artefacts in the reconstruction, 
a rigid monolithic map needs constant updates, 
which is often a computationally expensive process. 

Several state-of-the-art dense reconstruction systems have been developed in recent years to handle dynamic environments~\cite{Azim2012iv, Newcombe2015CVPR_dynamicfusion, Schmid2023dynablox, Schmid2024rss_Khronos}. 
DynamicFusion~\cite{Newcombe2015CVPR_dynamicfusion} proposed a dense warp field to transform non-rigid TSDF surfaces into a rigid canonical space, 
producing impressive reconstruction results on difficult non-rigid objects, such as humans. 
However, DynamicFusion tracks all potential dynamics in the observation and estimates the warp field for the whole canonical space, 
limiting this system to one small-scale non-rigid scene due to its computation complexity



For navigation in large-scale environments with many dynamic objects, 
Dynablox~\cite{Schmid2023dynablox} and Khronos~\cite{Schmid2024rss_Khronos} use TSDF and focus on accurately representing the free space by explicitly mapping the environment between the sensor and observed surfaces. 
The volume occupied by dynamic objects, 
on the other hand, 
are not modelled, 
as they are represented only using point clouds. 
Conversely, 
Fusion++~\cite{McCormac20193dv_fusionplusplus} and MID-Fusion~\cite{xu2019icra} volumetrically and incrementally reconstruct dynamic objects using TSDF, 
but do not explicitly represent all the free space from the sensor to object surfaces due to the truncation in TSDF. 
The work of Azim~\etal\cite{Azim2012iv} creates an \emph{occupancy} representation of the whole environment, and detects, segments and tracks moving objects in the scene. 
However, it does not accumulate the tracked observations of each dynamic object into a complete reconstruction and therefore is unable to accurately represent the total space occupied by the object.


Instead of using voxel maps, 
MaskFusion employs a surfel representation to efficiently and incrementally reconstruct dynamic objects~\cite{Runz2018ismar_maskfusion}. 
Surfels are well suited to accommodate scene dynamics as they can be moved around and updated with ease~\cite{Whelan2015rss_elasticfusion, Runz2018ismar_maskfusion}. 
Alternatively, the work of Barad~\etal\cite{Barad2024arxiv} incrementally models dynamic objects using a 3D Gaussian Splatting (3DGS) framework, 
and can efficiently update the reconstruction once any movement is observed. 
However, both methods omit the known free space, requiring additional heuristics and strategies for path planning applications~\cite{Funk2021ral_supereight, Wang2021ras}. 
We therefore employ voxel maps in DynORecon to explicitly map the known free space.  
DynORecon models the environment using ESDF because it provides a more accurate representation than TSDF~\cite{Oleynikova2017iros_voxblox}. 
Instead of computing ESDF from TSDF (as in \cite{Oleynikova2017iros_voxblox} and \cite{reijgwart2020ral_voxgraph}), 
we leverage VDB-GPDF~\cite{wu2024vdb_gpdf} to directly estimate ESDF.

In recent years, 
submapping techniques have been employed by several dense reconstruction and SLAM systems,
inspired by works such as~\cite{Bosse2003}. 
Replacing one monolithic map with a collection of submaps allows incorporating SLAM loop closure corrections to the global reconstruction~\cite{Kahler2016eccv_infinitam, Whelan2015rss_elasticfusion, reijgwart2020ral_voxgraph, Wang2021ras}. 
DynORecon represents each dynamic object using a similar submap structure but offers a fundamentally new perspective on updating and maintaining each submap by using object motions estimates from a Dynamic SLAM pipeline. The use of submaps allows us to 
efficiently update our dynamic object reconstruction within a globally consistent reconstruction, ensures that these objects do not lead to artefacts in the static map.

\section{Method}
\label{sec:method}


DynORecon reconstructs the static environment and the dynamic objects in it with the following inputs: 
a trajectory of the sensor (e.g. RGB-D camera or LiDAR), 
labelled 3D point clouds indicating the static background and tracked dynamic objects, as well as the motions of observed objects. 
This information is commonly provided by most state-of-the-art Dynamic SLAM systems~\cite{bescos2021ral, judd2024ijrr_mvo, morris2024icra}, 
with which DynORecon is designed to run in parallel. 
\algref{alg:system_arch} provides an overview on the system architecture. 


\begin{algorithm}[t]
	\caption{ \small{System Architecture.}}
	\label{alg:system_arch}
 \small{
    \SetAlgoLined
    \DontPrintSemicolon
    \textbf{input}: New point measurements $\mathcal{M}_k$ labelled by object IDs $j \in \{1\cdots N\}$, object motions $\objmotion{\worldf}{k-1}{k}^j$, sensor pose $\campose{\worldf}{k}$,\;
    \textbf{output}: A static map $\mathcal{R}^0$ and $N$ object submaps $\mathcal{R}^j$\;
    \Begin{
        Separate Static Cloud $\mathcal{M}^0_k$ from Object Clouds $\mathcal{M}^j_k$\;
        Integrate $\mathcal{M}^0_k$ using $\campose{\worldf}{k}$ into $\mathcal{R}^0$ (\secref{sec:integration})\;
        \For {Each Object Cloud $\mathcal{M}^j_k$ (\secref{sec:local_recon})}
        {
            Find free space based on $\mathcal{M}^j_k$ and $\campose{\worldf}{k}$\;
            \uIf{New object}
            {   
                Compute initial object pose $\objpose{\worldf}{k}^j$\;
                Create new object submap $\mathcal{R}^j$\;
                Integrate $\mathcal{M}^j_k$ into $\mathcal{R}^j$ based on $\objpose{\worldf}{k}^j$\;
            } \Else 
            {   
                Propagate $\objpose{\worldf}{k-1}^j$ with $\objmotion{\worldf}{k-1}{k}$\;
                Integrate $\mathcal{M}^j_k$ into $\mathcal{R}^j$ based on $\objpose{\worldf}{k}^j$\;
            }
        }
    }
 }
\end{algorithm} 

\subsection{Notations and Background}

Three main reference frames are used in our system. 
The world frame $\{\worldf\}$ defines the fixed global reference frame. 
The sensor $\{\camf_k\}$ and object $\{\objf_k\}$ frames are associated with the sensor and object poses $\campose{\worldf}{k}, \objpose{\worldf}{k} \in \SE$, represented in $\{\worldf\}$ at discrete time-step $k$. 

For each 3D point measurement in the sensor frame $\nhpoint{\camf_k}{k}$, $\mpoint{\camf_k}{k} = \left[\nhpoint{\camf_k}{k}, 1\right]^\top$ defines its homogeneous coordinates. 
$\mathcal{M}_k$ is a set of any points at time-step $k$ where $\mpoint{}{k} \in \mathcal{M}_k$.
We omit the time-step notation $k$ for any time-independent variables, and are therefore static within their own reference frame.
Consequently,  
a static point in the world frame is \mbox{$\mpoint{\worldf}{} = \campose{\worldf}{k} \: \mpoint{\camf_k}{k}$}, 
and any point on a rigid-body represented in the object frame, $\{\objf_k\}$ is $\mpoint{\objf}{}$.

DynORecon assumes that moving objects are rigid bodies. 
Previous works~\cite{Chirikjian17idetc, morris2024icra} demonstrate that, 
under this assumption, 
there exists a single $\SE$ motion in $\{\worldf\}$ that propagates all rigid body map points from $\mpoint{\worldf}{k-1}$ to $\mpoint{\worldf}{k}$:
\begin{align}
    \label{eq:motion_point_rigid}
    \mpoint{\worldf}{k} =& 
    \objmotion{\worldf}{k-1}{k} \: \mpoint{\worldf}{k-1} \\
    \label{eq:motion_kinematic_constraint}
    \objpose{W}{k} =& \objmotion{W}{k-1}{k} \: \objpose{W}{k-1}\text{,}
\end{align}
where $\mpoint{\worldf}{k-1}$ and $\mpoint{\worldf}{k}$ is the same point on the object at consecutive time-steps. 
Vitally, our representation of $\objmotion{\worldf}{}{}$ preserves the structure of rigid bodies. 
By expressing a dynamic object point at different time-steps in the object's frame of reference:
\begin{equation}
\begin{aligned}
\mpoint{\worldf}{k} &= \objpose{\worldf}{k} \: \mpoint{\objf_k}{k} \\
\mpoint{\worldf}{k-1} &= \objpose{\worldf}{k-1} \: \mpoint{\objf_k}{k-1}\text{,}
\label{equ:object_point_in_local}    
\end{aligned}
\end{equation}
we can verify that motion model in~\eqref{eq:motion_kinematic_constraint} keeps $\mpoint{\objf}{}$ static within the object local frame by substituting~\eqref{equ:object_point_in_local} into~\eqref{eq:motion_point_rigid}:
\begin{equation}
\begin{aligned}
    \label{eq:motion_rigid_prove}
    \objpose{\worldf}{k} \:\mpoint{\objf_k}{k} &= \objmotion{\worldf}{k-1}{k} \: \objpose{\worldf}{k-1} \:\mpoint{\objf_{k-1}}{k-1} \\
     \objpose{\worldf}{k} \: \mpoint{\objf_k}{k} &= \objpose{\worldf}{k} \: \mpoint{\objf_{k-1}}{k-1} \\
      \mpoint{\objf_k}{k} &= \mpoint{\objf_{k-1}}{k-1} 
\end{aligned}
\end{equation}

The per-object motion $ \objmotion{\worldf}{k-1}{k}$, 
estimated directly by our Dynamic SLAM framework~\cite{morris2024icra},
anchors moving object points represented in the local frame $\{ \objf \}$ on each rigid-body to a global map in the world frame $\{ \worldf \}$. 
This ensures that individual maps remain consistent with the global map. Importantly, using $\objmotion{\worldf}{k-1}{k}$ allows us to arbitrary define the object pose, $\objpose{\worldf}{}$, without any prior information of the object model.
The specific representation of motion in the world frame used here can be computed using other common Dynamic SLAM outputs such as object poses~\cite{Chirikjian17idetc, morris2024icra}




\subsection{Scan Integration}
\label{sec:integration}

\begin{figure}[t]
	\centering
	\includegraphics[trim={0cm 0cm 0cm 0cm},clip,width=0.9\columnwidth]{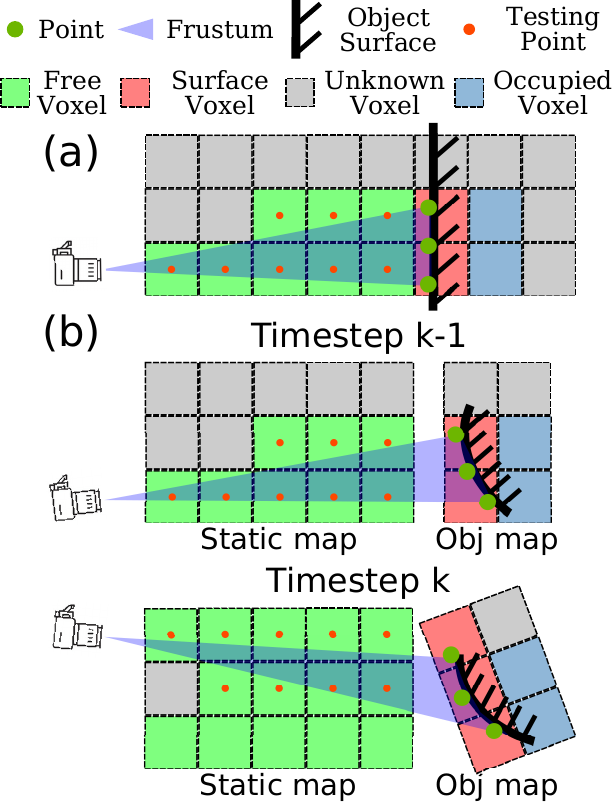}
	\caption{\small{A 2D example of ESDF update via VDB-GPDF for \textbf{(a)} static scene and \textbf{(b)} dynamic object. 
    Green points represent surface measurements and inform the surface voxels. 
    Within the sensor frustum, we sample testing points (orange) to update known free voxels. 
    Within a truncation distance from the surface along surface normals, voxels have their ESDF updated as occupied voxels.}}
    \label{fig:occupancy_update_example}
    \vspace{-6mm}
\end{figure}

We adopt scan integration and data fusion methods from the VDB-GPDF mapping framework~\cite{wu2024vdb_gpdf} to provide efficient updates and direct access to ESDF. 
Firstly, we group the raw measurements $\mathcal{M}_k$ at time-step $k$ into a set of voxels via a local VDB. 
We then sample between the sensor and the surfaces to obtain a set of testing voxels $\mathcal{V}_k$, 
efficiently representing the known free space without the need to update all voxels within the frustum. 
Using voxels contained in each leaf node of the local VDB as training points, 
we estimate a temporary latent Local Gaussian Process Signed Distance Field (L-GPDF), 
allowing us to query an ESDF distance and their uncertainties for every voxel in the sensor's frustum~\cite{wu2024vdb_gpdf}.
The local VDB is then fused into the gloabl VDB structure via a weighted sum update. 
Given the location of a testing voxel $\mathbf{v} \in \mathcal{V}_k$, 
we query the L-GPDF for the value of the distance field $d_k(\mathbf{v})$ and its variance $\sigma_k(\mathbf{v})$. 
The previous ESDF distance mean and weight are obtained from the global VDB, denoted as $\mu_{k-1}(\mathbf{v})$ and $\omega_{k-1}(\mathbf{v})$ respectively. 
We then perform distance fusion following the standard weighted sum as in~\cite{tsdf_fusion1996} to estimate the latest $\mu_{k}(\mathbf{v})$ and $\omega_{k}(\mathbf{v})$:
\begin{equation}
\begin{aligned}
\label{eq:weighted_sum}
\mu_k(\mathbf{v}) & =\frac{\omega_{k-1}(\mathbf{v}) \cdot \mu_{k-1}(\mathbf{v})+(1-\sigma_k(\mathbf{v})) \cdot d_k(\mathbf{v})}{\omega_{k-1}(\mathbf{v})+(1-\sigma_k(\mathbf{v}))} \\
\omega_k(\mathbf{v}) & =\omega_{k-1}(\mathbf{v})+(1-\sigma_k(\mathbf{v})).
\end{aligned}
\end{equation}
After integration, the dense reconstruction is updated by applying Marching Cubes~\cite{lorensen1998marching} to all voxels that participated in the most recent fusion process.

\subsection{Map Structure}
\label{sec:map_structure}

In traditional dense reconstruction methods, which create a static and rigid representation, dynamic objects move relative to both the global map and their current reconstruction. 
As a result, the entire structure of the object must be updated to account for its new pose.
This approach is computationally inefficient and can lead to residual artefacts in the static regions of the map as it is difficult to ensure that only the object model is updated without accurate tracking. 

Instead of creating one monolithic global reconstruction, 
DynORecon maintains a collection of small-scale submaps corresponding to each dynamic object in addition to one global map, covering the static background. 
This strategy is used in~\cite{Kahler2016eccv_infinitam, xu2019icra, Wang2021ras} to efficiently handle relative movements among maps. 
We employ this technique to facilitate highly efficient object reconstruction updates. 



DynORecon volumetrically represents the object using voxels in each object submap, 
as indicated by~\figref{fig:occupancy_update_example}. 
For each object we build our reconstruction around its body-fixed frame $\objf_k$ that is associated with object pose $\objpose{\worldf}{k}$.
Within this fixed frame, the object reconstruction and the point measurements of the object surface are time invariant, 
allowing new observations to be integrated into the reconstruction without updating the entire object. 
Furthermore, representing the object as a time-invariant reconstructions allows us to accommodate object motion without a computationally heavy voxel map update.
As the object moves, 
we update the pose of the object relative to the static map using the motion model~\eqref{eq:motion_kinematic_constraint}, 
ensuring the object frame moves consistently with the physical object and any point measurements on it.

\subsection{Object Local Reconstruction}
\label{sec:local_recon}

Different from the conventional scan integration strategy discussed in~\secref{sec:integration}, 
we integrate the points on dynamic objects into their individual submaps. 
DynORecon first conducts the process of scan integration as explained in~\secref{sec:integration} using point measurements in the world frame $\mpoint{\worldf}{k}$. 
However, we process the occupied and free voxels separately as demonstrated in~\figref{fig:occupancy_update_example}. 

Observed free voxels are integrated into the global static map to better facilitate navigation tasks.
Occupied voxels on dynamic objects on the other hand, are not updated in the static map, 
but in their corresponding object submaps instead.
To represent each point in the submaps fixed frame, DynORecon transforms observed dynamic points into their local frame $\{ \objf \} $ using \eqref{equ:object_point_in_local}.



By representing each submap as locally static,
we can model object motion without updating the whole object reconstruction, 
but simply via propagating the object pose $\objpose{\worldf}{}$ every frame using the motion model in~\eqref{eq:motion_point_rigid} and the estimated object motion $\objmotion{\worldf}{}{}$.
This propagation moves the reference frame $\{\objf_k\}$ of the object reconstruction consistently with its corresponding object points as proven in~\eqref{eq:motion_rigid_prove}, 
facilitating a rigid object-centric reconstruction. 
The uniqueness of this submap representation capitalises on the motion estimation $\objmotion{\worldf}{}{}$ provided by our Dynamic SLAM pipeline~\cite{morris2024icra} to efficiently and accurately represent the dynamics of each object. 

Given no prior information, 
we assign the fixed reference frame of each object upon its first observation with an initial pose, 
which is defined using the centroid of the initial 3D point measurements as its position and the identity matrix as its orientation.
Although this assignment is arbitrary (e.g., for a car, the x-axis may not align with the direction of motion), 
the orientation of the reference frame is irrelevant because $\objmotion{\worldf}{k-1}{k}$ will correctly propagate all points on the rigid-body regardless of the orientation of the reference frame. 
This also applies to the reference frame's position, 
as the initial centroid represents a point on the rigid body that can be propagated through accurate motion estimation in the world frame, 
and does not need to be located at any specific position on the object, i.e. the actual centre of mass.



\subsection{ESDF Query}

For the purpose of exploration and navigation, 
DynORecon uses the surface information provided by ESDF to represent occupied space. 
However, as each object is represented by its own reconstruction,
querying the ESDF of any 3D point involves a time consuming search through all relevant maps.
In order to improve the computation efficiency, 
we first check the desired coordinate with the bounding box of each map to only proceed with ESDF query in near-by maps.
After gathering the ESDF of the desired coordinate, 
we apply the same weighted sum operation as~\eqref{eq:weighted_sum} in~\secref{sec:integration}, 
to obtain the final ESDF value, 
and in turn represent obstacles. 



\section{Experiments}

We tested our proposed system with a series of experiments using data from both simulation and the real world to demonstrate its performance, i.e. incrementally reconstructing dynamic objects and explicitly modelling the free space. 


\begin{figure}[t]
	\centering
	\includegraphics[trim={0cm 0cm 0cm 0cm},clip,width=0.9\columnwidth]{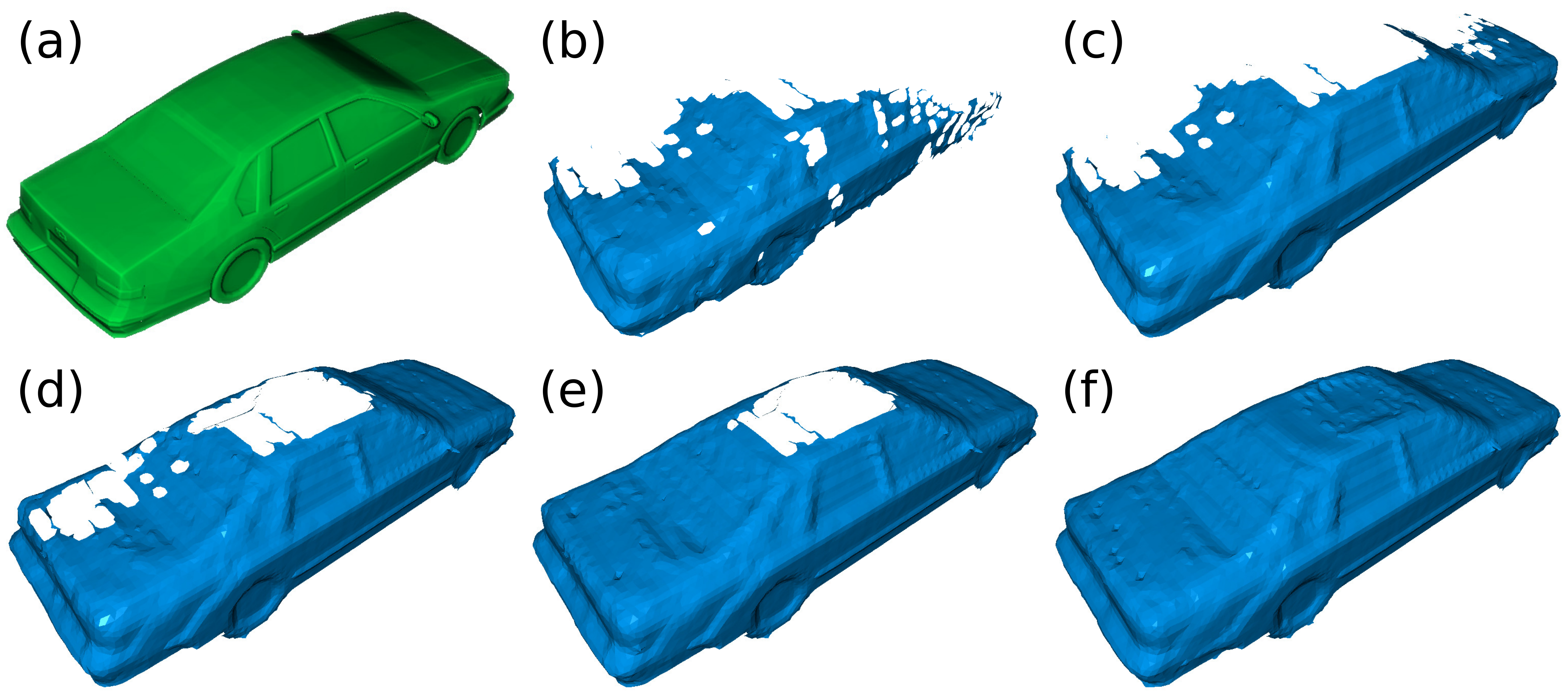}
	\caption{\small{Incremental reconstruction results in the Simulation experiment. \textbf{(a)} - ground truth mesh; \textbf{(b-f)} - the reconstruction of the vehicle is accumulated consistently with more observations.}}
    \label{fig:simulation_results}
    \vspace{-6mm}
\end{figure}

\begin{figure}[b]
    \vspace{-4mm}
	\centering
	\includegraphics[trim={0cm 0cm 0cm 0cm},clip,width=0.9\columnwidth]{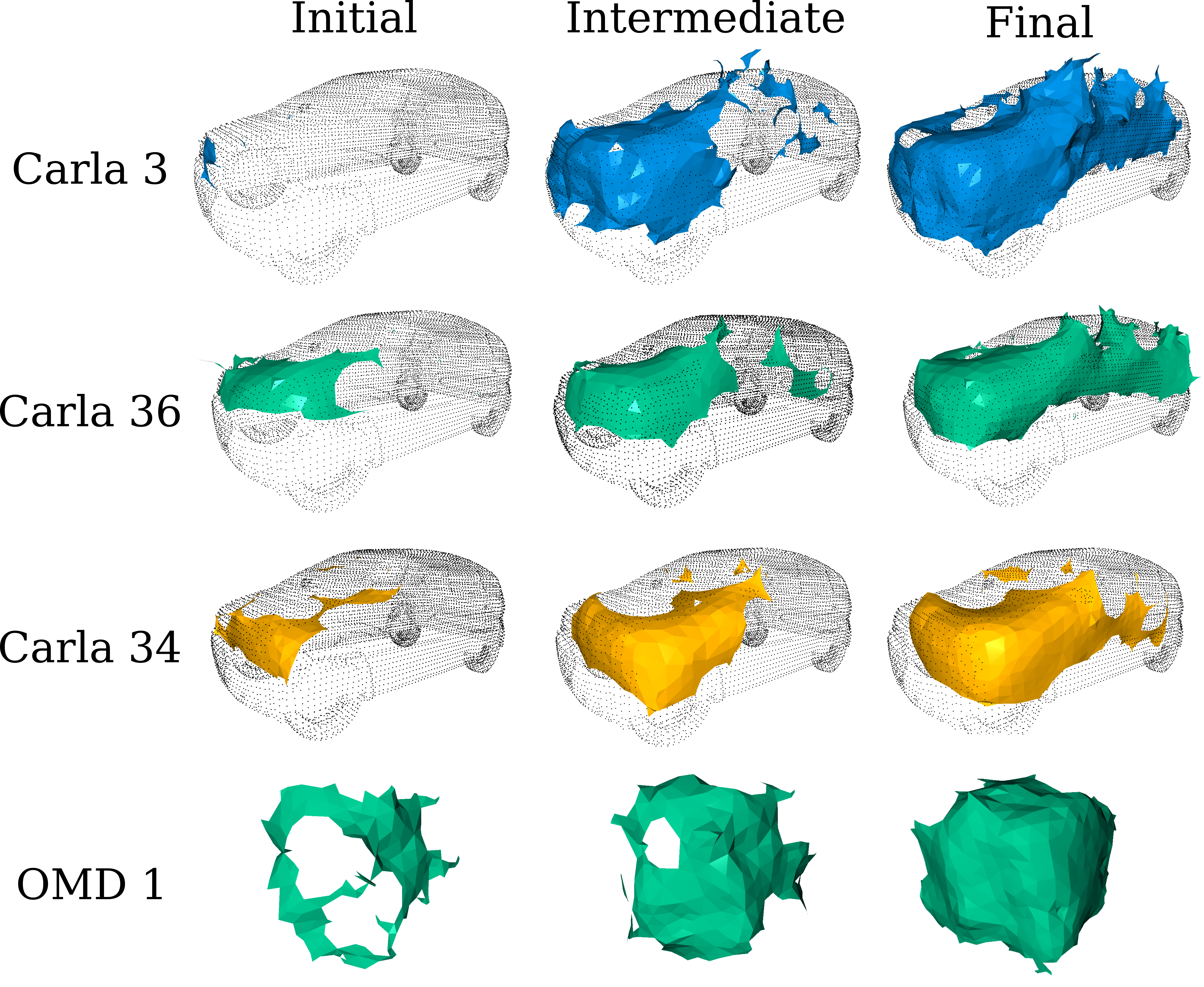}
	\caption{\small{Incremental reconstruction results on 3 cars from the Outdoor Cluster dataset~\cite{Huang2019iccv} and a cube in OMD~\cite{Judd19ral}.
    The incremental reconstructions are shown at three different phases. 
    Outdoor Cluster reconstructions are aligned with the ground truth point cloud (shown in black) for visual comparison.}}
    \label{fig:results}
\end{figure}



To evaluate DynORecon, we utilized four different datasets: Outdoor Cluster~\cite{Huang2019iccv}, Oxford Multimotion Dataset (OMD)~\cite{Judd19ral}, DOALS~\cite{Pfreundschuh2021icra_doals} and \emph{Simulated}. 
The Outdoor Cluster dataset from ClusterSLAM~\cite{Huang2019iccv} was used to assess the coverage and accuracy of our reconstruction, as it provides ground truth object models. 
For OMD,
we selected the most challenging sequence, \emph{swinging 4 unconstrained} to test our pipeline with complex object movement. This dataset features four swinging cubes with unconstrained and unpredictable motion, and was used for qualitative evaluations. 
The DOALS dataset was used to evaluate DynORecon's free space mapping accuracy. Finally, we created a simulated sequence (\textit{Simulation}) to verify DynORecon's ability to fully map a moving object. 

The voxel resolution was set to \SI{10}{\centi\meter} for the Outdoor Cluster and Simulation experiments, and \SI{5}{\centi\meter} for the OMD experiment due to its small-scale indoor environment. Both datasets were pre-processed using the Dynamic SLAM system from our previous work~\cite{morris2024icra} to estimate camera poses, object motions, and provide labeled 3D point clouds.

\subsection{Incremental Reconstruction of Dynamic Objects}

To evaluate DynORecon's ability to incrementally reconstruct dynamic objects, we measure the percentage coverage of the reconstructed mesh from Marching Cubes~\cite{lorensen1998marching} by comparing it to the ground truth model.
The coverage percentage is computed as: 
\begin{equation}
    r_\text{cov} =  \frac{N_\text{covered gt}}{N_\text{all gt}}\text{,}
\end{equation}
where $N_\text{all gt}$ is the number of all points in the ground truth map, 
and $N_\text{covered gt}$ is the number of points in the ground truth that has a neighbouring point in the reconstruction with a distance $d < \lambda_{cover}$. 
In this experiment, $\lambda_{cover} = 0.05\sqrt{3}$~\SI{}{\meter}, 
which is the distance from the center of a \SI{10}{\centi\meter} voxel to its vertex. 

Based on CARLA simulator, we designed the \emph{Simulation} experiment, 
where we model a flying camera that scans a moving vehicle from all angles (see~\figref{fig:simulation_results} (a)). 
We used perfectly known trajectories and noiseless measurements to eliminate any imperfections when we evaluate \emph{coverage}. 
The greedy scanning pattern ensures that the object is fully observed, unlike other tested datasets where only partial views of dynamic objects are available due to the trajectories of the objects relative to the camera. 

\begin{figure}[b]
    \vspace{-6mm}
	\centering
	\includegraphics[trim={0cm 0cm 0cm 0cm},clip,width=\columnwidth]{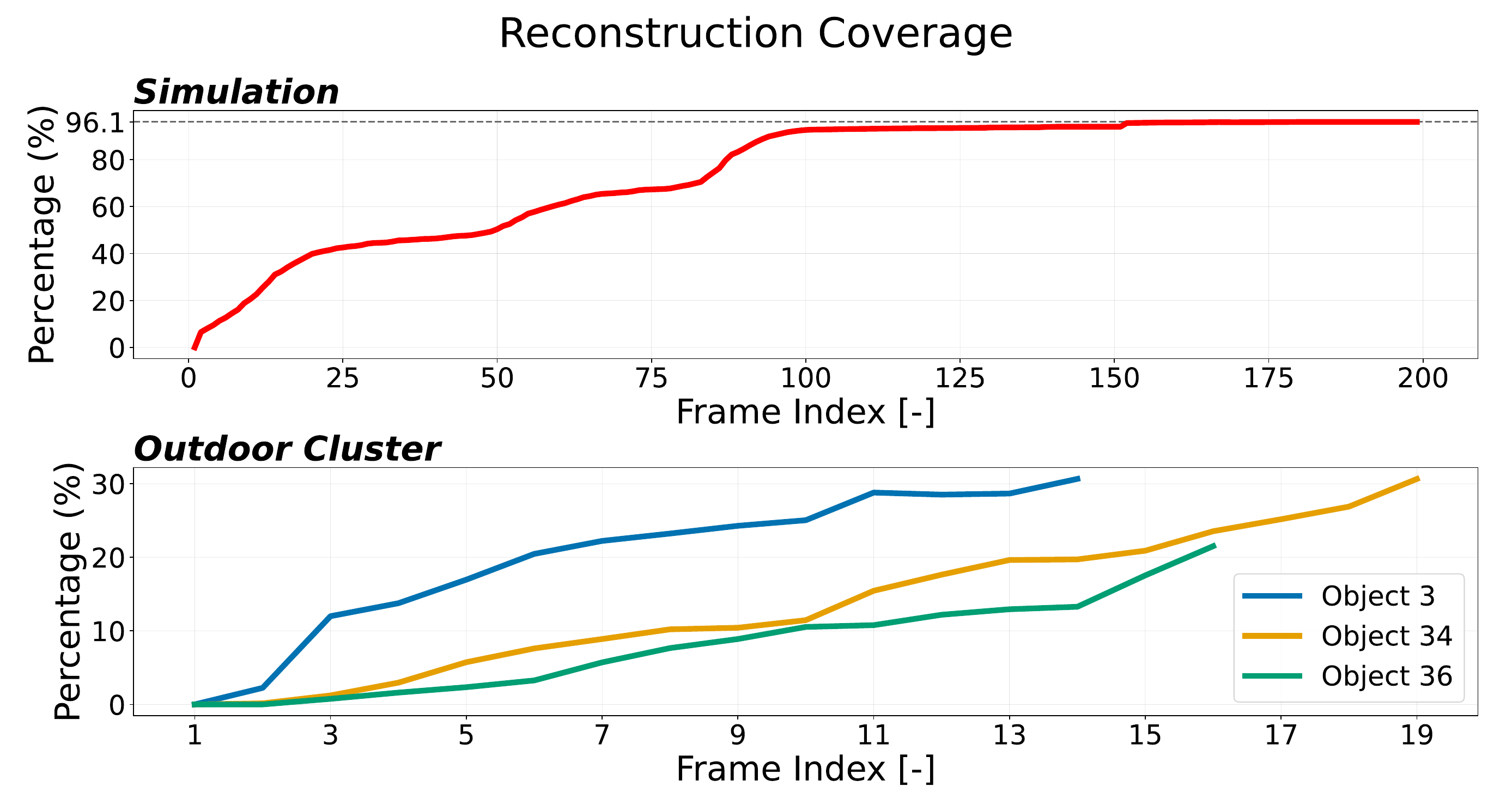}
	\caption{\small{Per-frame reconstruction percentage coverage on the \textit{Simulation} (top) and \textit{Outdoor Cluster} (bottom) datasets. The frame index shows the number of frames from the first observation until the object goes out of view.}}
    \label{fig:combined_coverage_plots}
\end{figure}

Since we incrementally build our reconstruction, 
we are able to evaluate the coverage at every time-step against the ground truth model as more of the object is observed by the sensor. 
The per-frame coverage for our simulation is shown in~\figref{fig:combined_coverage_plots} (top) which clearly demonstrates the growth in coverage, achieving $96.1\%$ in the end. 
\figref{fig:simulation_results} (b-f) further presents a visual demonstration of the incremental reconstruction. 
The remaining uncovered surfaces are mostly from the underside of the tyres and the loss of fine surface details from the \SI{10}{\centi\meter} mesh resolution. 



\begin{table}[t]
\footnotesize
\centering
\setlength{\tabcolsep}{4.8pt}
\caption{\small{RMSE of final object reconstruction using the Outdoor Cluster dataset.}}
\label{tab:carla_rmse}
\begin{tabular}{c|cccc}
\toprule
 Object & 3 (car) & 36 (car) & 34 (car) \\
\midrule
RMSE (m) & $0.12$ & $0.17$ & $0.14$ \\
\bottomrule
\end{tabular}
\vspace{-6mm}
\end{table}

 We then tested DynORecon using the Outdoor Cluster experiment, a simulated outdoor sequence.
\figref{fig:results} presents the close-up incremental reconstruction results of a vehicles object at different time-steps, and \figref{fig:combined_coverage_plots} (bottom) presents the growth in the reconstruction of the same object.

In this experiment, 
only the left side of the objects are reconstructed as these vehicles are visible for a short period of time and are commonly driving towards the sensor on the left-side of the road, 
limiting our potential reconstruction coverage. 
However, DynORecon successfully builds up their reconstructions with only a few observations. 
\tabref{tab:carla_rmse} presents the accuracy of the reconstructed meshes compared with the ground truth models. We calculate the error by measuring the distance from each vertex in the mesh to the ground truth and reported the RMSE. The reconstruction error is approximately \SI{15}{\centi\meter}, only slightly exceeding the experiment's voxel resolution of \SI{10}{\centi\meter}, and comparable to the highly accurate results achieved by the Dynamic SLAM~\cite{morris2024icra}.

\figref{fig:results} also presents one of the four cubes in the OMD experiment~\cite{Judd19ral} being mapped on-the-fly as both the camera and the objects underwent unpredictable motions. 
Despite the complex motion of the cube in the OMD dataset,
our pipeline is able to qualitatively reconstruct the cube after only a few frames. 
Our results show that DynORecon is able to reconstruct dynamic objects from accumulated observations with high accuracy and consistency. 

\subsection{Computation Efficiency}
The computation time of DynORecon was evaluated in the Outdoor Cluster experiment covering a large-scale outdoor environment and utilizing a dense depth camera with a resolution of $1280\times720$. 
Because this experiment is in simulation, 
we choose to use as far as \SI{25}{\meter} maximum sensor range, 
mapping much more space than conventional dense cameras can manage. 
Despite these challenges, the average time required to integrate a new scan for a dynamic object is approximately \SI{10}{\milli\second}, while integration on new scans into the static map takes an average of \SI{45}{\milli\second}.
Such a performance is appropriate for real-time online applications 
that requires quick responses such as obstacle avoidance, 
and downsampling the measurements can further improve the computation time.

\subsection{Free Space}

\begin{table}[t]
\footnotesize
\centering
\setlength{\tabcolsep}{4.8pt}
\caption{\small{Number of free space voxels reconstructed compared to ground truth using DOALS dataset~\cite{Pfreundschuh2021icra_doals}: HC - High Confidence; LC - Low Confidence; TP - True Positive; FN - False Negative; FP - False Positive.}}
\label{tab:free_space}
\begin{tabular}{c|ccc|ccc}
\toprule
  & \multicolumn{3}{c|}{Hauptgebaeude} & \multicolumn{3}{c}{Station} \\
Ground Truth Free Space  & \multicolumn{3}{c|}{$5871$} & \multicolumn{3}{c}{$27548$} \\
\midrule
  & TP & FN & FP & TP & FN & FP \\
\midrule
Dynablox HC Free Space & $2790$ & $161$ & $65$ & $19288$ & $108$ & $125$ \\
Dynablox LC Free Space & $5639$ & $161$ & $65$ & $24384$ & $108$ & $125$ \\
Ours & $3980$ & $16$ & $93$ & $25214$ & $7$ & $148$\\
\bottomrule
\end{tabular}
\vspace{-6mm}
\end{table}

Given the importance of explicitly known free space in robotic navigation tasks, we evaluated DynORecon's \emph{free space} reconstruction performance using the DOALS dataset~\cite{Pfreundschuh2021icra_doals}. 
\tabref{tab:free_space} presents a comparison between the reconstructed free space against the ground truth using \textit{True Positive} and \textit{False Negative}. 
We accumulate the ground truth static point clouds provided by DOALS into a dense map, 
and greedily sample scan positions within the map to perform ray-casting and compute ground truth free space. 
To evaluate the free space, 
we iterate through all known free voxels within the ground truth map and check the corresponding voxel in the reconstruction. 
If both voxels are free, it is True Positive; 
if the reconstruction considers it occupied, it is False Negative. 
By iterating through occupied voxels, 
we further report the False Positive voxels, i.e. occupied in the ground truth but free in the reconstruction. 
\begin{figure}[h]
    \vspace{-3mm}
	\centering
	\includegraphics[trim={0cm 0cm 0cm 0cm},clip,width=0.98\columnwidth]{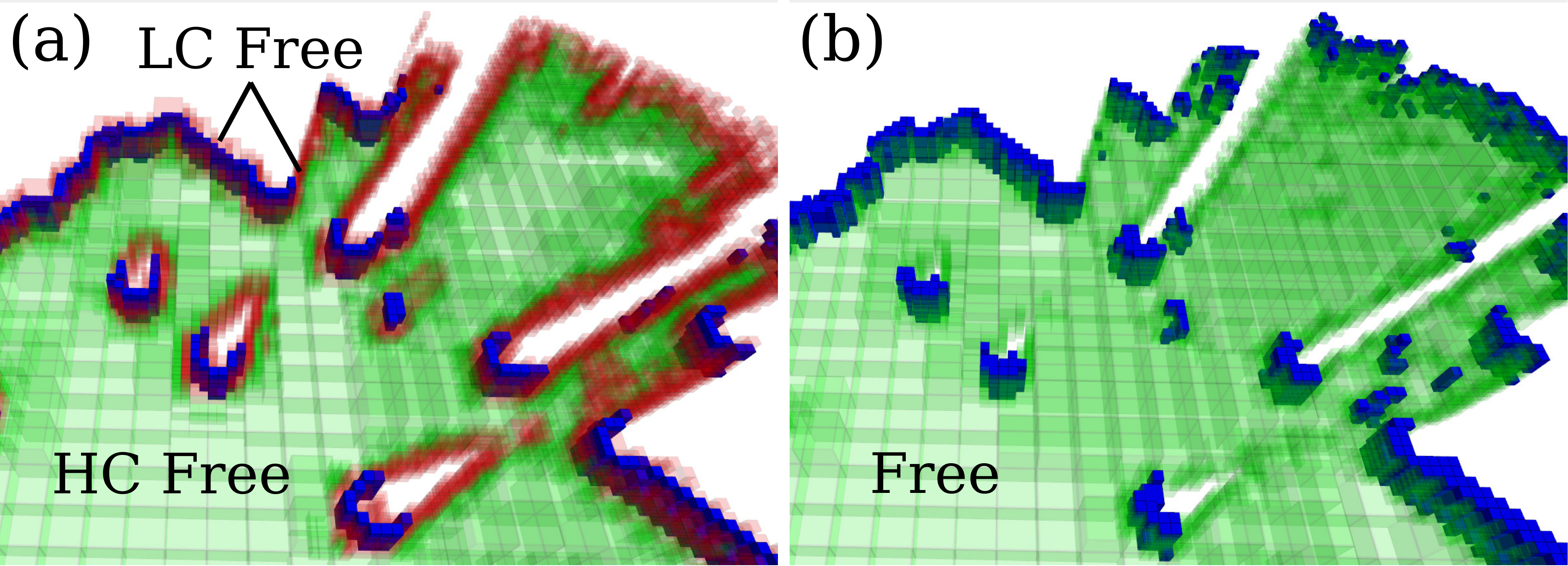}
	\caption{\small{A slice of reconstructions created by \textbf{(a)} Dynablox and \textbf{(b)} DynORecon in the Hauptgebaeude sequence. We visualise the free space using translucent green voxels and occupied space using solid blue voxels, and low confidence free space of Dynablox is further highlighted using translucent red voxels in \textbf{(a)}.}}
    \label{fig:free_space}
    \vspace{-3mm}
\end{figure}
The evaluation results are reported in~\tabref{tab:free_space} for both DynORecon and state-of-the-art system Dynablox~\cite{Schmid2023dynablox}. 
Dynablox focuses on free space reconstruction in dynamic environments and directly segments out dynamic points from LiDAR scans.
On the other hand, 
DynORecon is coupled with a visual SLAM system~\cite{morris2024icra} that leverages image semantics and motions to generate a segmented point cloud. 
For a fair comparison, 
we use the point cloud segmented by Dynablox and the same voxel resolution (\SI{20}{\centi\meter}), truncation distance (\SI{40}{\centi\meter}) and maximum sensor range (\SI{20}{\meter}) to reconstruct the free space map.
In addition, we take a \SI{1}{\meter} slice centred around the height of the LiDAR from all the reconstructions because the greedily sampled free space in the ground truth map exceeds the maximum scanning range of the LiDAR.

As demonstrated by~\tabref{tab:free_space}, 
both systems are capable of detecting the majority of the free space with occlusion and limited sensor range being the main cause of unobserved free space, 
and both systems accurately map the occupied space with very few False Positive around the static scene surfaces. 
Dynablox considers the free space around object surfaces and the boundary of known and unknown space as low confidence free space,  
while the remaining free space as high confidence as visualised in~\figref{fig:free_space}. 
Consequently, 
DynORecon reconstructs more free space than the high confidence free space mapped by Dynablox,
while the low confidence free space of Dynablox reports more True Positive results. 
However, some of the additional free space is in the unknown space, such as behind observed surfaces (\figref{fig:free_space}). 
DynORecon detects more free space in the Station sequence where the LiDAR explored an open space with little to no static occlusions. 
Furthermore, 
Dynablox conservatively marks more free voxels as occupied (False Negative) around object surfaces compared to DynORecon, 
which can become a hindrance for path planning algorithms through tight corridors. 
While Dynablox is shown highly effective in detecting dynamic points using free space, 
we believe that DynORecon is more appropriate for navigation. 


\section{Conclusion and Future Work} 
\label{sec:conclusion}

In summary, we present DynORecon, 
a reconstruction system for dynamic environments with a focus on incrementally mapping dynamic objects. 
The proposed system further maintains an explicit representation of observed free space for navigation and obstacle avoidance purposes. 
For dynamic objects, 
we process observed occupied and free space using different strategies so as to avoid residual artefacts due to their movements. 

For further development, 
we are first interested in handling non-rigid-body objects, which are abundant in natural environments. 
Our intended strategies include tracking each rigid segments separately, 
or incorporating a more complicated motion model as inspired by BodySLAM++~\cite{Henning2023iros_bodyslam++}. 
For deformable objects, 
we would like to explore a parameterised multi-motion model or a deformable function, which will be a direct extension to our existing formulation. 


Currently, our system relies on the upstream SLAM component to 
track 3D points with consistent object labels and motions based on a rigid-body motion assumption between consecutive frames.
We plan to further leverage the volumetric map to filter out 3D point outliers in the Dynamic SLAM pipeline, thereby improving the reconstruction in future versions of DynORecon. 
We are also interested in incorporating the voxel map directly into the Dynamic SLAM framework, 
computing cost functions directly on the dense reconstruction. 






\bibliographystyle{IEEEtran}
\bibliography{./IEEEabrv, ./refs/bibliography, ./refs/reference_vdb}

\begin{thebibliography}{10}
\providecommand{\url}[1]{#1}
\csname url@rmstyle\endcsname
\providecommand{\newblock}{\relax}
\providecommand{\bibinfo}[2]{#2}
\providecommand\BIBentrySTDinterwordspacing{\spaceskip=0pt\relax}
\providecommand\BIBentryALTinterwordstretchfactor{4}
\providecommand\BIBentryALTinterwordspacing{\spaceskip=\fontdimen2\font plus
\BIBentryALTinterwordstretchfactor\fontdimen3\font minus
  \fontdimen4\font\relax}
\providecommand\BIBforeignlanguage[2]{{%
\expandafter\ifx\csname l@#1\endcsname\relax
\typeout{** WARNING: IEEEtran.bst: No hyphenation pattern has been}%
\typeout{** loaded for the language `#1'. Using the pattern for}%
\typeout{** the default language instead.}%
\else
\language=\csname l@#1\endcsname
\fi
#2}}

\bibitem{hornung13ar_octomap}
A.~Hornung, K.~M. Wurm, M.~Bennewitz, C.~Stachniss, and W.~Burgard,
  ``{OctoMap}: An efficient probabilistic {3D} mapping framework based on
  octrees,'' \emph{Autonomous Robots}, 2013.

\bibitem{Niessner2013tog}
M.~Nie{\ss}ner, M.~Zollh\"ofer, S.~Izadi, and M.~Stamminger, ``Real-time 3d
  reconstruction at scale using voxel hashing,'' \emph{ACM Transactions on
  Graphics (TOG)}, 2013.

\bibitem{Kahler2015tvcg_infinitam}
O.~Kahler, V.~A. Prisacariu, C.~Y. Ren, X.~Sun, P.~H.~S. Torr, and D.~W.
  Murray, ``{Very High Frame Rate Volumetric Integration of Depth Images on
  Mobile Device},'' \emph{IEEE Trans. on Visualization and Computer Graphics},
  vol.~22, no.~11, 2015.

\bibitem{bescos2021ral}
B.~Bescos, C.~Campos, J.~D. Tardós, and J.~Neira, ``Dynaslam ii:
  Tightly-coupled multi-object tracking and slam,'' \emph{{IEEE} Robotics and
  Automation Letters}, vol.~6, no.~3, pp. 5191--5198, 2021.

\bibitem{morris2024icra}
J.~Morris, Y.~Wang, and V.~Ila, ``The importance of coordinate frames in
  dynamic slam,'' in \emph{Proc. of the IEEE Intl. Conf. on Robotics and
  Automation (ICRA)}, 2024.

\bibitem{judd2024ijrr_mvo}
K.~M. Judd and J.~D. Gammell, ``Multimotion {Visual} {Odometry} ({MVO}),''
  \emph{Intl. J. of Robotics Research}, 2024.

\bibitem{Funk2021ral_supereight}
N.~Funk, J.~Tarrio, S.~Papatheodorou, M.~Popović, P.~F. Alcantarilla, and
  S.~Leutenegger, ``Multi-resolution 3d mapping with explicit free space
  representation for fast and accurate mobile robot motion planning,''
  \emph{{IEEE} Robotics and Automation Letters}, vol.~6, no.~2, pp. 3553--3560,
  2021.

\bibitem{reijgwart2020ral_voxgraph}
V.~{Reijgwart}, A.~{Millane}, H.~{Oleynikova}, R.~{Siegwart}, C.~{Cadena}, and
  J.~{Nieto}, ``Voxgraph: Globally consistent, volumetric mapping using signed
  distance function submaps,'' \emph{{IEEE} Robotics and Automation Letters},
  vol.~5, no.~1, pp. 227--234, 2020.

\bibitem{Wang2021ras}
Y.~Wang, M.~Ramezani, M.~Mattamala, S.~T. Digumarti, and M.~Fallon,
  ``Strategies for large scale elastic and semantic lidar reconstruction,''
  \emph{J. of Robotics and Autonomous Systems}, vol. 155, 2022.

\bibitem{Newcombe2015CVPR_dynamicfusion}
R.~A. Newcombe, D.~Fox, and S.~M. Seitz, ``Dynamicfusion: Reconstruction and
  tracking of non-rigid scenes in real-time,'' in \emph{Proc. of the IEEE Intl.
  Conf. Computer Vision and Pattern Recognition}, June 2015.

\bibitem{Runz2018ismar_maskfusion}
M.~Runz, M.~Buffier, and L.~Agapito, ``Maskfusion: Real-time recognition,
  tracking and reconstruction of multiple moving objects,'' in \emph{IEEE/ACM
  Intl. Sym. on Mixed and Augmented Reality (ISMAR)}, 2018, pp. 10--20.

\bibitem{Schmid2023dynablox}
L.~Schmid, O.~Andersson, A.~Sulser, P.~Pfreundschuh, and R.~Siegwart,
  ``Dynablox: Real-time detection of diverse dynamic objects in complex
  environments,'' \emph{{IEEE} Robotics and Automation Letters}, vol.~8,
  no.~10, pp. 6259 -- 6266, 2023.

\bibitem{Barad2024arxiv}
K.~R. Barad, A.~Richard, J.~Dentler, M.~Olivares-Mendez, and C.~Martinez,
  ``Object-centric reconstruction and tracking of dynamic unknown objects using
  3d gaussian splatting,'' 2024.

\bibitem{Schmid2024rss_Khronos}
L.~Schmid, M.~Abate, Y.~Chang, and L.~Carlone, ``Khronos: A unified approach
  for spatio-temporal metric-semantic slam in dynamic environments,'' in
  \emph{Proc. of the Robotics: Science and Systems (RSS)}, 2024.

\bibitem{Oleynikova2017iros_voxblox}
H.~Oleynikova, Z.~Taylor, M.~Fehr, R.~Siegwart, and J.~Nieto, ``Voxblox:
  Incremental 3d euclidean signed distance fields for on-board mav planning,''
  in \emph{Proc. of the IEEE/RSJ Intl. Conf. on Intelligent Robots and Systems
  (IROS)}, 2017, pp. 1366--1373.

\bibitem{wu2024vdb_gpdf}
L.~Wu, C.~L. Gentil, and T.~Vidal-Calleja, ``Vdb-gpdf: Online gaussian process
  distance field with vdb structure,'' \emph{arXiv preprint arXiv:2407.09649},
  2024.

\bibitem{wu2021faithful}
L.~Wu, K.~M.~B. Lee, L.~Liu, and T.~Vidal-Calleja, ``Faithful euclidean
  distance field from log-gaussian process implicit surfaces,'' \emph{IEEE
  Robotics and Automation Letters}, vol.~6, no.~2, pp. 2461--2468, 2021.

\bibitem{legentil2023accurate}
C.~Le~Gentil, O.-L. Ouabi, L.~Wu, C.~Pradalier, and T.~Vidal-Calleja,
  ``Accurate gaussian-process-based distance fields with applications to
  echolocation and mapping,'' \emph{IEEE Robotics and Automation Letters},
  vol.~9, no.~2, pp. 1365--1372, 2024.

\bibitem{lorensen1998marching}
W.~E. Lorensen and H.~E. Cline, ``Marching cubes: A high resolution 3d surface
  construction algorithm,'' in \emph{Seminal graphics: pioneering efforts that
  shaped the field}, 1998, pp. 347--353.

\bibitem{museth2013vdb}
K.~Museth, ``Vdb: High-resolution sparse volumes with dynamic topology,''
  \emph{ACM transactions on graphics (TOG)}, vol.~32, no.~3, pp. 1--22, 2013.

\bibitem{museth2013openvdb}
K.~Museth, J.~Lait, J.~Johanson, J.~Budsberg, R.~Henderson, M.~Alden, P.~Cucka,
  D.~Hill, and A.~Pearce, ``Openvdb: an open-source data structure and toolkit
  for high-resolution volumes,'' in \emph{Acm siggraph 2013 courses}, 2013, pp.
  1--1.

\bibitem{Newcombe2011ismar}
R.~A. Newcombe, S.~Izadi, O.~Hilliges, D.~Molyneaux, D.~Kim, A.~J. Davison,
  P.~Kohi, J.~Shotton, S.~Hodges, and A.~Fitzgibbon, ``Kinectfusion: Real-time
  dense surface mapping and tracking,'' in \emph{IEEE/ACM Intl. Sym. on Mixed
  and Augmented Reality (ISMAR)}, 2011, pp. 127--136.

\bibitem{Azim2012iv}
A.~Azim and O.~Aycard, ``Detection, classification and tracking of moving
  objects in a 3d environment,'' in \emph{{IEEE} Intelligence Vehicles Symp.
  (IV)}, 2012, pp. 802--807.

\bibitem{McCormac20193dv_fusionplusplus}
J.~Mccormac, R.~Clark, M.~Bloesch, A.~Davison, and S.~Leutenegger, ``Fusion++:
  Volumetric object-level slam,'' in \emph{Proc. of Intl. Conf. on 3D Vision},
  2018, pp. 32--41.

\bibitem{xu2019icra}
B.~Xu, W.~Li, D.~Tzoumanikas, M.~Bloesch, A.~Davison, and S.~Leutenegger,
  ``Mid-fusion: Octree-based object-level multi-instance dynamic slam,'' in
  \emph{Proc. of the IEEE Intl. Conf. on Robotics and Automation (ICRA)}.\hskip
  1em plus 0.5em minus 0.4em\relax IEEE, 2019, pp. 5231--5237.

\bibitem{Whelan2015rss_elasticfusion}
T.~Whelan, S.~Leutenegger, R.~Salas-Moreno, B.~Glocker, and A.~Davison,
  ``Elasticfusion: Dense slam without a pose graph,'' in \emph{Proc. of the
  Robotics: Science and Systems (RSS)}, 2015.

\bibitem{Bosse2003}
M.~{Bosse}, P.~{Newman}, J.~{Leonard}, M.~{Soika}, W.~{Feiten}, and
  S.~{Teller}, ``An {Atlas} framework for scalable mapping,'' in \emph{Proc. of
  the IEEE Intl. Conf. on Robotics and Automation (ICRA)}, vol.~2, 2003, pp.
  1899--1906 vol.2.

\bibitem{Kahler2016eccv_infinitam}
O.~K{\"{a}}hler, V.~A. Prisacariu, and D.~W. Murray, ``Real-time large-scale
  dense 3d reconstruction with loop closure,'' in \emph{Proc. of the European
  Conf. on Computer Vision (ECCV)}, 2016, pp. 500--516.

\bibitem{Chirikjian17idetc}
G.~S. Chirikjian, R.~Mahony, S.~Ruan, and J.~Trumpf, ``Pose changes from a
  different point of view,'' in \emph{Proc. of the ASME Intl. Design
  Engineering Technical Conf. (IDETC)}.\hskip 1em plus 0.5em minus 0.4em\relax
  ASME, 2017.

\bibitem{tsdf_fusion1996}
B.~Curless and M.~Levoy, ``A volumetric method for building complex models from
  range images,'' in \emph{Proceedings of the 23rd Annual Conference on
  Computer Graphics and Interactive Techniques}, ser. SIGGRAPH '96.\hskip 1em
  plus 0.5em minus 0.4em\relax Association for Computing Machinery, 1996, p.
  303–312.

\bibitem{Huang2019iccv}
J.~{Huang}, S.~{Yang}, Z.~{Zhao}, Y.~{Lai}, and S.~{Hu}, ``Clusterslam: A slam
  backend for simultaneous rigid body clustering and motion estimation,'' in
  \emph{Proc. of the Intl. Conf. on Computer Vision (ICCV)}, 2019, pp.
  5874--5883.

\bibitem{Judd19ral}
K.~M. Judd and J.~D. Gammell, ``{The Oxford Multimotion Dataset: Multiple SE(3)
  Motions with Ground Truth},'' \emph{{IEEE} Robotics and Automation Letters},
  vol.~4, no.~2, pp. 800--807, 2019.

\bibitem{Pfreundschuh2021icra_doals}
P.~Pfreundschuh, H.~F. Hendrikx, V.~Reijgwart, R.~Dubé, R.~Siegwart, and
  A.~Cramariuc, ``Dynamic object aware lidar slam based on automatic generation
  of training data,'' in \emph{Proc. of the IEEE Intl. Conf. on Robotics and
  Automation (ICRA)}, 2021, pp. 11\,641--11\,647.

\bibitem{Henning2023iros_bodyslam++}
D.~F. Henning, C.~Choi, S.~Schaefer, and S.~Leutenegger, ``Bodyslam++: Fast and
  tightly-coupled visual-inertial camera and human motion tracking,'' in
  \emph{Proc. of the IEEE/RSJ Intl. Conf. on Intelligent Robots and Systems
  (IROS)}, 2023, pp. 3781--3788.

\end{thebibliography}

\end{document}